\documentclass[web]{ieeecolor}
\usepackage{sub}
\usepackage{cite}
\usepackage{amsmath,amssymb,amsfonts}
\usepackage{algorithm2e}
\usepackage{graphicx}
\usepackage{textcomp}
\usepackage{array}
\usepackage{graphicx}
\usepackage{caption}
\usepackage{subcaption}
\usepackage{enumerate}
\usepackage{booktabs}
\usepackage{multirow}
\usepackage{threeparttable}
\usepackage{verbatim}
\usepackage{placeins}
\usepackage{flushend}
\usepackage{bbm}
\newcommand{\myparagraph}[1]{\vspace{0.1em}\noindent\textbf{#1}}
\usepackage[colorlinks,hyperindex,breaklinks]{hyperref}
\usepackage[noend]{algpseudocode}
\usepackage[font=small,labelfont=bf,justification=justified,format=plain]{caption} 
\newcommand{\ie}{\textit{i}.\textit{e}.}
\newcommand{\eg}{\textit{e}.\textit{g}.}
\newcommand{\etal}{\textit{et al}.}
\newlength\myindent
\def\BibTeX{{\rm B\kern-.05em{\sc i\kern-.025em b}\kern-.08em
		T\kern-.1667em\lower.7ex\hbox{E}\kern-.125emX}}
\begin{document}
         \title{Non-parametric Regularization for Class Imbalance Federated Medical Image Classification}
	\author{Jeffry Wicaksana, Zengqiang Yan, and Kwang-Ting Cheng, \IEEEmembership{Fellow, IEEE}
		\thanks{
			Jeffry Wicaksana and Kwang-Ting Cheng are with the Department of Electronic and Computer Engineering, Hong Kong University of Science and Technology, Kowloon, Hong Kong (E-mail: jwicaksana@connect.ust.hk, timcheng@ust.hk).
		}
		\thanks{
			Zengqiang Yan is with the School of Electronic Information and Communications, Huazhong University of Science and Technology, Wuhan, China (E-mail: z\_yan@hust.edu.cn).
		}
	}
	\maketitle
	
	\begin{abstract} 
        Limited training data and severe class imbalance pose significant challenges to developing clinically robust deep learning models. 
        Federated learning (FL) addresses the former by enabling different medical clients to collaboratively train a deep model without sharing privacy-sensitive data. However, class imbalance worsens due to variation in inter-client class distribution. 
        We propose federated learning with non-parametric regularization (FedNPR and FedNPR-Per, a personalized version of FedNPR) to regularize the feature extractor and enhance useful and discriminative signal in the feature space. Our extensive experiments show that FedNPR outperform the existing state-of-the art FL approaches in class imbalance skin lesion classification and intracranial hemorrhage identification. 
        Additionally, the non-parametric regularization module consistently improves the performance of existing state-of-the-art FL approaches.
        We believe that NPR is a valuable tool in FL under clinical settings. 
        The code is available at:
        https://github.com/Jwicaksana/FedNPR.
	\end{abstract}
	
	\section{Introduction} 
	\label{sec:introduction}
	
	\IEEEPARstart{C}{omputer}-aided diagnosis based on medical image content analysis is a valuable tool for assisting professionals in decision making and patient screening. Deep learning models have achieved impressive success in various tasks, including skin lesion classification~\cite{skin1,skin2} of dermoscopy images, intracranial hemorrhage~\cite{rsna} identification of CT images, and autism disorder prediction~\cite{sheller,sheller2} of FMRI images. However, training a robust model requires a large amount of annotated data, which is often infeasible to collect without infringing patient's privacy. 
	
    Federated learning (FL) enables different medical clients to collaboratively train a federated model~\cite{fedavg,sheller,sheller2,dou_npj} in a privacy-preserving manner. In FL, a server facilitates collaboration by exchanging model weights instead of patients' data. A federated training round consists of two stages: 1) \textbf{local update}, where each client downloads the federated model from the server and updates it locally, and 2) \textbf{server update}, where the server aggregates model updates from each client and updates the federated model. The training repeats until convergence. 

\begin{figure}[t!]
        \centering
        \includegraphics[width=1 \columnwidth]{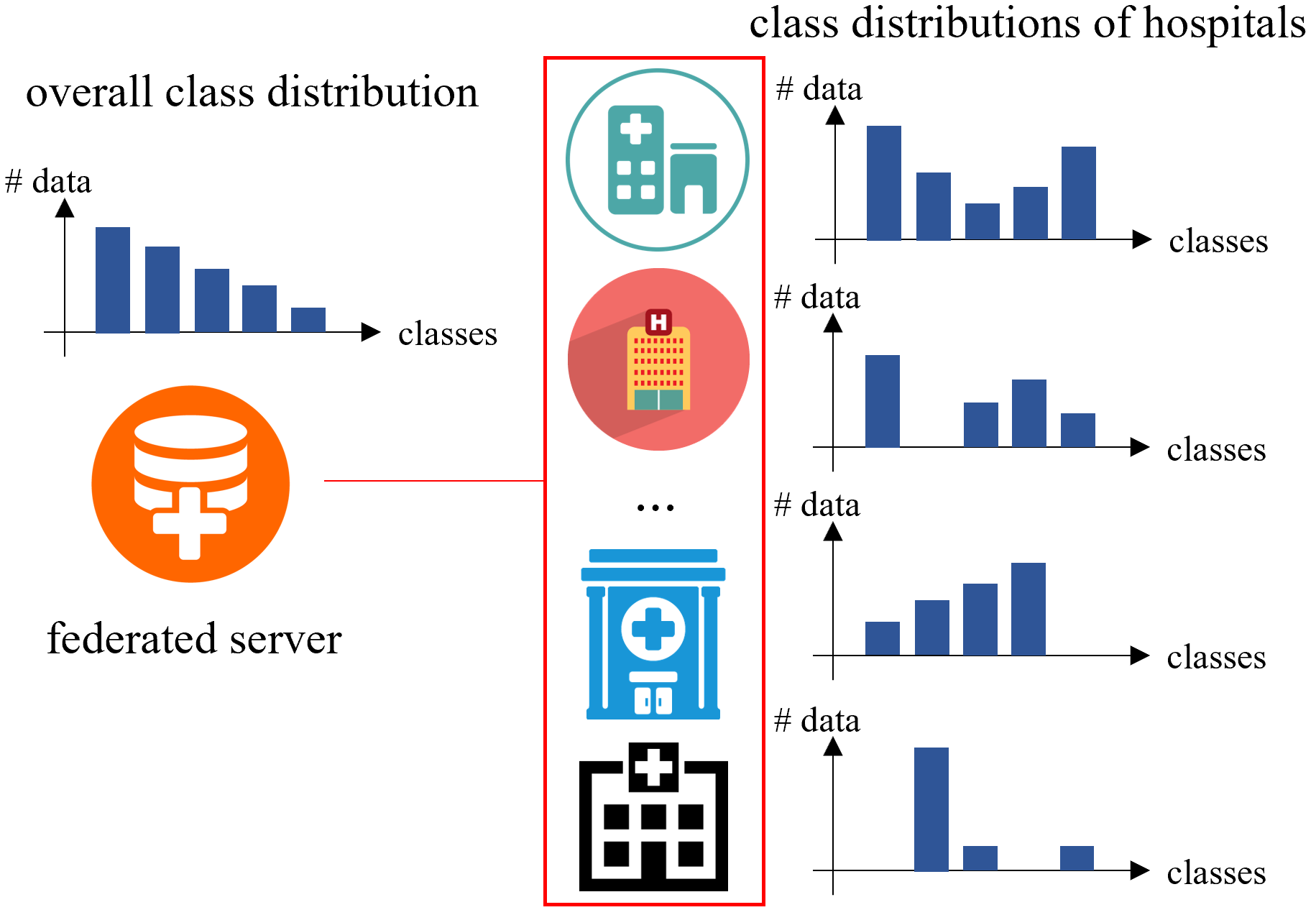}
        \caption{Variations of class distributions under the federated setting between: 1) different clients, where some clients have missing and rare classes, and 2) each individual client and the global federation.} \label{class_statistics}
    \end{figure}

    Under the federated setting, learning robust deep models for medical applications is challenging as
    medical data is imbalance in nature~\cite{ham10k,isic2017,bcn2000}. Federated class imbalance is not only present within each client's data but also across clients, as shown in Fig.~\ref{class_statistics}. 
    Limited data on certain classes hinders data-driven/parametric approach to learn a robust feature extractor. Consequently, local model updates from various clients may diverge, leading to unstable update~\cite{fedprox} and sub-optimal model performance~\cite{silo}.
    
    We propose a non-parametric regularization (NPR) module to enhance expresiveness and discriminativeness of the extracted features. With more generalizable feature extractor, local update from different clients are more consistent, leading to a more robust and accurate of the federated model. 
    Inspired by the recent successes on using feature prototype~\cite{prototype} in few-shot and zero-shot learning, NPR explicitly models and injects class-specific characteristics into the feature space to regularize the feature extractor. We use multiple class prototypes~\cite{rss_prototype,dnc} to capture better class-specific characteristics.
    The feature prototypes are obtained through online sub-clustering with sinkhorn iteration~\cite{sinkhorn}. 
    Additionally, NPR can be seamlessly integrated with various existing federated learning (FL) techniques. In this work, we demonstrate how to integrate (NPR) into Federated Learning (FL) to optimize a global federated model, referred to as FedNPR, and to optimize multiple personalized models, denoted as FedNPR-Per. 
    
    Existing federated approaches handle inter-client class variations through: 1) single model regularization, or 2) multiple models personalization. Single model regularization imposes constraints to each client by regularizing the model update with respect to the global model~\cite{fedprox,moon} or global class distribution~\cite{balancefl}, while personalization allows different clients to globally share a part of the model, \eg, the feature extractor, and learn their own client-specific classification head~\cite{pros_nas,prr,cusfl,fedrep,fedbabu}. Both approaches overlook the fact that each client often has limited data, which can result in an inadequate learning of robust feature extractors during the local update. Under severe class imbalance and inter-client class variations, this can lead to extracted features that are not representative or reliable, ultimately resulting in a sub-optimal federated model.

    Comprehensive evaluation across real-world class imbalance federated datasets for skin lesion \cite{ham10k,isic2017,bcn2000} and intracranial hemorrhages classification~\cite{rsna} demonstrates that: 1) FedNPR and FedNPR-Per exhibit superior performance compared to state-of-the-art FL approaches, and 2) NPR can be effectively integrated with various FL techniques to enhance model performance. 
    
    Our contributions are:
    \begin{itemize}
        \item We introduce a non-parametric regularization (NPR) module for class imbalance federated learning. NPR enhances the robustness of the extracted features and enriches features diversity by explicitly modeling class-relevant information in each client. Additionally, NPR is designed to be compatible with existing FL techniques, as presented in Sections~\ref{framework}. 
        \item Building upon the non-parametric regularization (NPR) framework, we propose two novel federated learning (FL) techniques: FedNPR, a single-model regularization FL method, and FedNPR-Per, a personalized FL approach. The details of these techniques are presented in Sections~\ref{sec3:7}. 
        \item In Sections~\ref{evaluation}, we demonstrate that FedNPR and FedNPR-Per outperform state-of-the-art single-model and personalized FL techniques, respectively, on class imbalance real-life federated skin lesion classification and intracranial hemorrhage identification. Furthermore, we show in Section~\ref{discussion} that NPR consistently enhances the performance of existing FL methods. The relevant details of existing FL methods are provided in Sections~\ref{lit}.
    \end{itemize}

    \section{Related Work}\label{lit}

    \begin{figure*}[t!]
		\centering	\includegraphics[width=1\textwidth]{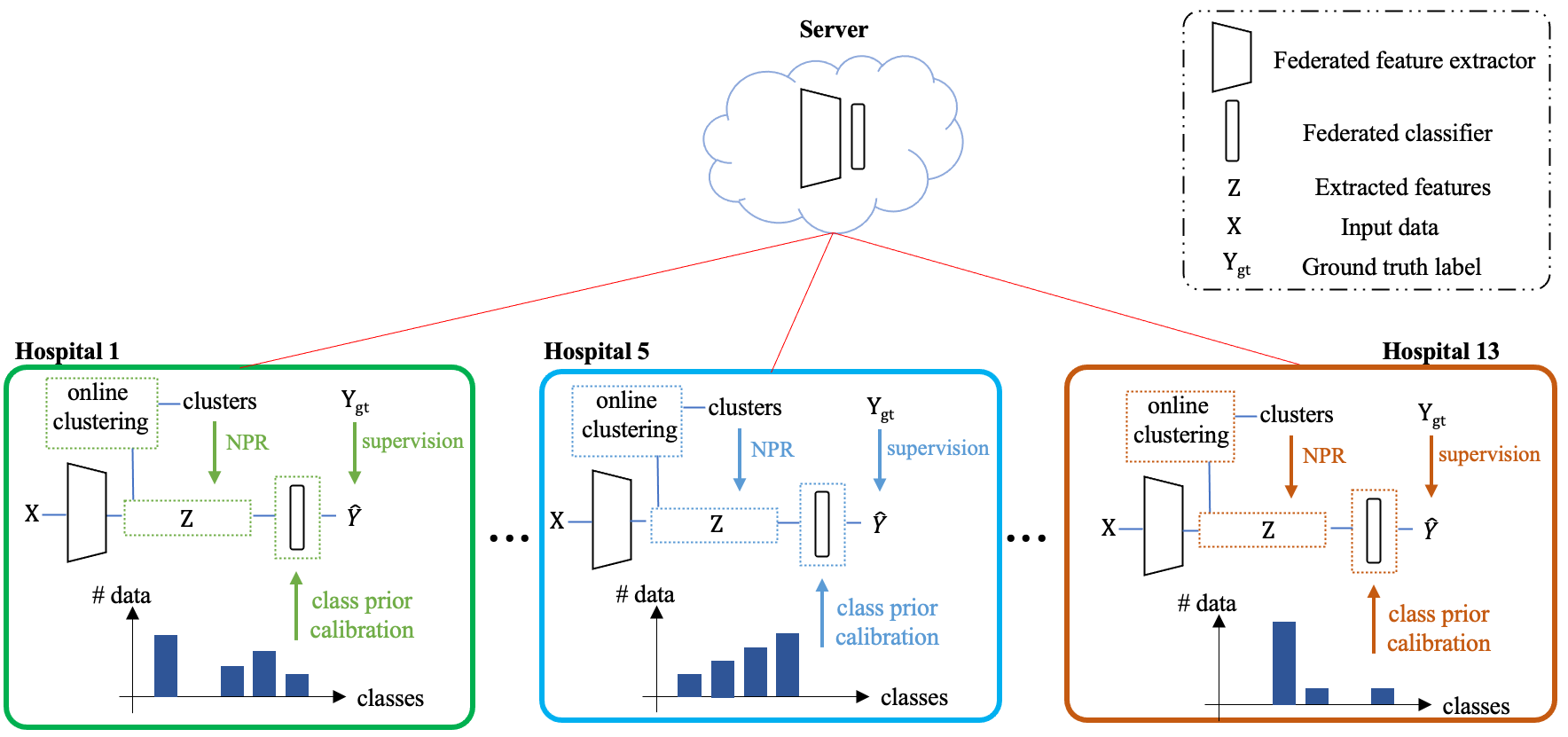}
		\caption{
			Illustration of federated learning with non-parametric regularization (FedNPR). FedNPR has two components: 1) non-parametric regularization, where each client first cluster its features into a K $\times$ $\#$ class clusters, which serve as anchors to regularize the extracted features, and 2) class-prior calibration according to each client's local class distribution. The regularization components are applied during the local update process, ensuring that the learned features capture the underlying data distribution.
		}
		\label{overview}
	\end{figure*}
    
    \subsection{Federated Learning}
    Federated learning (FL)~\cite{fedavg} has gained significant attention in the medical imaging communities~\cite{fl_natural, dou_npj} 
    due to its ability to enable multiple clients to collaboratively train a deep model without sharing raw data.
    This privacy-preserving technique has been applied to various multi-clients medical imaging applications, including functional magnetic resonance imaging classification~\cite{abide}, health tracking through wearables~\cite{fl_wearable}, COVID-19 screening and lesion detection~\cite{fl_covid}, brain tumor segmentation~\cite{fl_pp_brain, sheller,sheller2}, skin tumor classification~\cite{prr}, among others. 
	
    In real-world scenarios, each client collects data locally under different conditions and protocols~\cite{fl_dichotomy}, leading to inter-client statistical variations and non-independent and identically distributed (non-IID) data.
    To learn a robust model, it is essential to address the inter-client statistical heterogeneity~\cite{flamby,fedavg,flc}.
    Existing approaches for handling non-IID data can be broadly categorized into two types: single-model/global and multi-models/personalization.
	
    Single-model federated learning (FL) approaches aim to reduce the divergence between the global federated model and the local model updates of each client. FedProx~\cite{fedprox} and FedMA~\cite{fedma} minimize the divergence in model weights using a proximal term and a layer-wise model construction, respectively. FedDG~\cite{feddg}, VAFL~\cite{vafl}, and FedRobust~\cite{fedrobust} minimize variations in the sample space by mapping each client's data into a common domain. In the feature space, FTL~\cite{eegdomain} regularizes features across clients with a covariance matrix, MOON~\cite{moon} and FedCON~\cite{fedcon} use contrastive learning and FedBN~\cite{fedbn} leverages local batch normalization. CCVR~\cite{ccvr} argues that classifiers diverge the most and thus debiased each client's classifier with virtually generated features. 
	   
    Personalized FL resolves inter-client variation by locally isolating each client's unique characteristic. FedRep~\cite{fedrep} and CusFL~\cite{cusfl} trained a federated feature extractor and allowed each client to maintain its personalized classifier head. FedBABU~\cite{fedbabu} first freezes the classifier during training and then fine-tune it locally. FedRod~\cite{fedrod} enables participating clients to train its own classifier on top of the shared federated head. FedMD~\cite{fedmd} and PRR~\cite{prr} trains distinct network architectures for each client with knowledge transfer, while~\cite{pros_nas} utilized neural architecture search. 
    
    \subsection{class imbalance learning}
    In centralized learning, class imbalance is addressed through: 1) data re-sampling, 2) data re-weighting, 3) enhancing representation learning, or 4) multi-expert learning. Data re-sampling oversamples low-frequency classes~\cite{relay, calib_lt} or undersamples high-frequency classes~\cite{rethink_lt,systematic_imbalance}. Focal~\cite{focal_loss} and class-balanced loss~\cite{cbl} adjust the weight of each sample according to the training losses' magnitude and class frequencies respectively. Approaches focusing on enhancing representation increase class margins~\cite{ldam}, 
    utilize class prototypical embedding~\cite{paco}, calibrating classifiers to be class-balanced~\cite{balanced_softmax}, or train model in two stages~\cite{decouple}, \eg, representation learning and classifier learning. Multi-expert learning~\cite{bbn,trustworthy,sade} trains multiple classifier heads while sharing a feature extractor to handle different distributions.
	
    Federated class imbalance learning is more challenging than centralized class imbalance learning due to inter-client class variations. RatioLoss~\cite{fl_ci_sharing} estimates and reweights each class' importance by monitoring model gradients in the server using auxiliary data. Astraea~\cite{astra} introduces an oracle as to rebalance each client's class distribution. CReRF~\cite{crerf} recalibrates the federated classifier in the server using synthesized balanced features. FedIRM~\cite{fedirm} and imFedSemi~\cite{dynamic_bank} resolves class imbalance in semi-supervised settings by sharing class relation matrices and highly confident unlabeled samples respectively. However, sharing additional information beyond model weight update is not desirable in medical domains due to the potential risk of privacy leakage~\cite{leakage}. 
	
    To avoid sharing additional information, CLIMB~\cite{climb}, BalanceFL~\cite{balancefl}, and FedLC~\cite{fedlc} learn balanced federated model by reweighting each client's importance during aggregation based on the empirical loss, balanced class sampling with self-entropy regularization, and logits calibration with pair-wise margins respectively. FedRS~\cite{fedrs} limits classifier updates when there are missing classes. However, conforming the federated model to a balanced class distribution may be detrimental for some clients, \eg, being sub-optimal compared to their locally-learned models.

    FedNPR focuses on the local update phase of each client, enabling them to learn a more consistent and robust model update using a more generalizable federated feature extractor. This approach uses local regularization which eliminates the need to share additional knowledge.
    
    \subsection{Non-parametric Learning}
    Recent advances in few-shot~\cite{prototype, fewshot-proto} and zero-shot~\cite{zero_1,zero_2} learning have demonstrated the benefits of non-parametric approaches, such as prototypical learning, for quickly learning and modeling class-specific characteristics without requiring additional training round.  
    
    There have been attempts to combine supervised learning with non-parametric learning for semantic segmentation~\cite{rss_prototype} and classification~\cite{dnc}, resulting in a improved performance and better explainability. 

    FedNPR builds upon non-parametric learning~\cite{rss_prototype, dnc} to directly model each client's class structure and information into multiple sub-clusters in the feature space. The sub-clusters provide explicit guidance to regularize and enhance the generalizability of the feature extractor, particularly for minority/rare classes. 

    \section{Methodology}\label{framework} 
    We first introduce the notations in Section~\ref{sec3:1} and then provide an overview of non-parametric regularization (NPR) module in Section~\ref{sec3:2}. In Section~\ref{sec3:3} and~\ref{sec3:4}, we describe how to incorporate NPR under the federated setting, \eg, FedNPR. 
    Section~\ref{sec3:7} describes how to extend FedNPR as a personalized FL, \eg, FedNPR-Per. 
	
    \subsection{Preliminaries: notation}\label{sec3:1}
    We denote a deep model for classification $\phi_{w}$ as a combination of feature extractor $f_{u}$ and classifier $g_{v}$, \eg, $\phi_{w} = \{f_{u}, g_{v}\}$. $\phi_{w}$ is optimized with $N$ clients' training data $D \triangleq \bigcup_{k} D_{N}$. 
    Here, $D_{i} = \{X_{i}, Y_{i}\}$ represents each client $i$'s training data, $X_{i}=[x_{1},...,x_{|X_{i}|}]$, and $Y=[y_{1},...,y_{|Y_{i}|}] \in \{1, ..., C\}$. 
    For brevity, we denote $D_{i}^{c}$ as the training data of client $i$ belonging to class $c$, where $|D^c_i|=\sum_{y_{i}\in D_{i}} \mathbbm{1}_{y_i==c}$  and $Z_{i}=f_{u}(X_{i})$ as the set of feature obtained from client $i$'s training images.  
    
    With non-parametric regularization (NPR), each client $i$ clusters its $Z_{i}$ into $K \times C$ sub-clusters. Feature belonging to the same class $c$ are assigned into $K$ sub-clusters. The set of the center of clusters, \eg, prototypes, is denoted as $P_{i} = [{p_{1}}_{1},... ,{p_{1}}_{k},... ,{p_{C}}_{1},... ,{p_{C}}_{K}]$.  
    
    \RestyleAlgo{ruled}
	\SetKwComment{Comment}{/* }{ */}
	\begin{algorithm}[t]
		\caption{Pseudocode of FedNPR}\label{alg}
		\SetKwInOut{Input}{input}
		\SetKwInOut{Output}{output}
		\SetKwInOut{Parameter}{parameter}
		\Input{$D$: Training data of N clients}
		\Parameter{$\lambda$: hyperparameter controlling the importance of NPR \\
                $K$ : number of sub-clusters per class \\
			$\alpha$ : learning rate \\
			$T$: federated training rounds\\}
		\Output{$w^{T}$: federated model's parameters}
            $w^{0}$ $\leftarrow$ \textbf{initialize}()\\
		\For{$t=1:T$}{
			${\overline{\nabla}\phi} = \{\}$\\
			\For{$i=1:N$}{
				$\phi_{w}$ $\leftarrow$ \textbf{Download}($w^{t-1}$) \\
				${Z_{i}}$ $\leftarrow$ $g_{v}^{t-1} ({X}_i)$ \\
                   $P_{i} \leftarrow$ \textbf{Cluster} ${Z_{i}}$ into $KC$ sub-clusters\\
				$\nabla {\phi}^{i}_{w}$ $\leftarrow$ \textbf{Update}($\phi_{w};\alpha,\lambda,P_{i},D_i$)\\
				$\overline{\nabla}\phi$.add($\nabla{\phi}^{i}_{w}$) \\
			}
			${w}^{t}$ $\leftarrow$ \textbf{Aggregate}(${\overline{\nabla}\phi}$, $w^{t-1}$)\\
		}
		\Return{$w^{T}$}
	\end{algorithm}
    
    \subsection{Overview}\label{sec3:2}
    In this section, we provide an overview of both the non-parametric regularization (NPR) module and FedNPR. 
    
    \myparagraph{Non-parametric regularization (NPR)}: NPR consists of two components:
    \begin{enumerate}
        \item \textbf{Online sub-clustering:} NPR groups the features of each class $c$ into $K$ sub-clusters to better capture the characteristics of different classes.
        At each training round, the centers of the sub-clusters are updated based on the most recent extracted features, allowing the model to adapt to changing data distributions.
        NPR employs a sinkhorn iteration-based clustering algorithm~\cite{sinkhorn}, which is a fast and effective method for cluster assignment.
        \item \textbf{Feature regularization via sub-clusters:} NPR leverages the center of each class' sub-clusters as anchors to pull the neighboring features together, enhancing the features' signal and information from various classes by directly modeling the relationship in the feature space. 
        This approach not only strengthens the features' expressiveness but also improves features' generalization by increasing the features' compactness and inter-class separability. 
    \end{enumerate}

    NPR is designed as a modular component that can be seamlessly integrated with various existing federated learning techniques, as its primary role is to regularize the feature extractor. 

    \myparagraph{FedNPR:} FedNPR mitigates inter-client class variation by incorporating two complementary techniques: 1) non-parametric regularization (NPR), which serve to regularize and guide the extracted features. 2) class-prior calibration to debias the extracted features on each client according to its local class distribution. 
    
    Each training round of FedNPR, described in Section~\ref{sec3:4}, consists of two stages: 1) \textit{local client update} where each client $i$ downloads the federated model $\phi_{w}$ from the server and updates $\phi_{w}$ with its local data, and 2) \textit{server update}, where the server updates the federated model with local updates from participating clients $\overline{\nabla}\phi= \{\nabla\phi_{w}^{1}, ..., \nabla\phi_{w}^{k} \}$. The pseudocode of FedNPR is provided in Algorithm~\ref{alg} and the overview of FedNPR is presented in Figure~\ref{overview}.

     \begin{figure}[t!]
        \centering
        \includegraphics[width=1 \columnwidth]{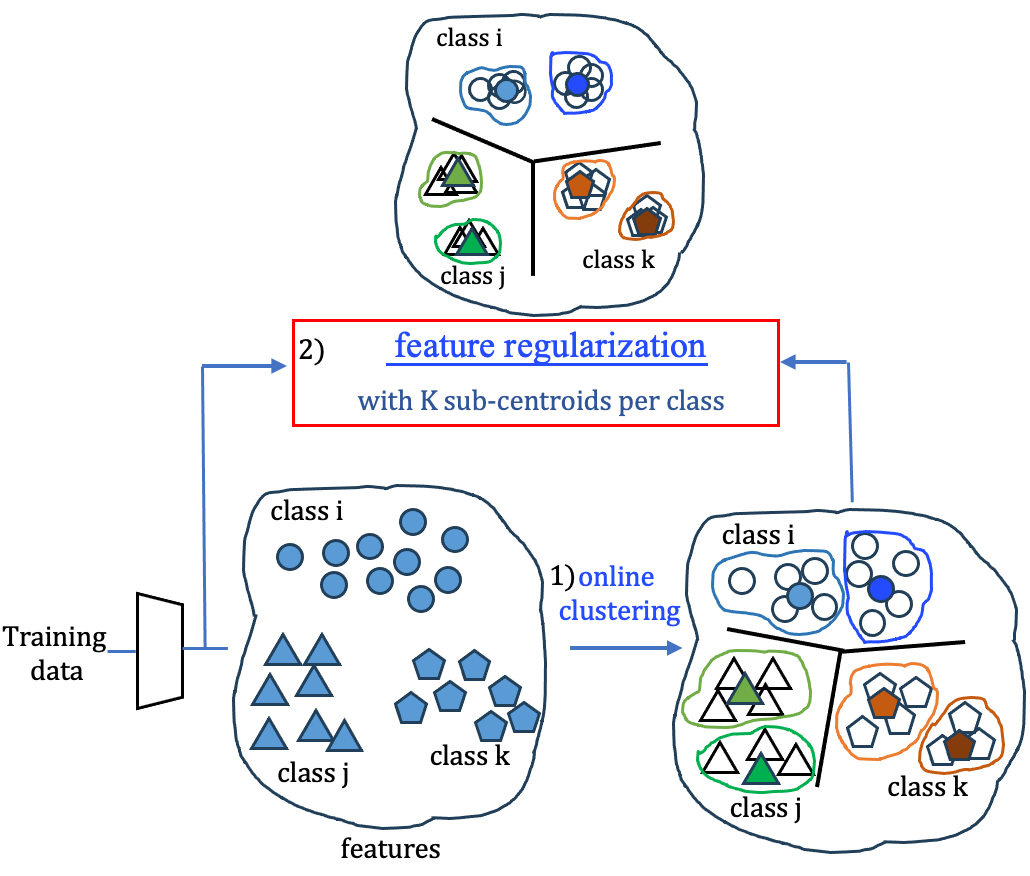}
        \caption{Illustration of the steps in non-parametric regularization (NPR) module: 
        1) online clustering of each client's extracted features into $KC$ sub-clusters with sinkhorn iteration~\cite{sinkhorn}, and 
        2) feature regularization using the center of the sub-clusters where we push the feature of each sample to be close to the corresponding sub-clusters and away from the others.}\label{npr_figure}
    \end{figure}

    \subsection{Non-parametric regularization}\label{sec3:3}
    Non-parametric regularization (NPR) has two components: 1) online sub-clustering and 2) feature regularization via sub-clusters. The overview of NPR is illustrated in Fig~\ref{npr_figure}

    \subsubsection{online sub-clustering} First, we find the mapping function $Q^{c} \in \mathbb{R}^{|D^{c}_{i}| \times K}$ that maps the $d$-dimensional normalized features $Z_{i}^{c} = Z_{i}^{c}/||Z_{i}^{c}||_{2} \in \mathbb{R}^{|D_{i}^{c}| \times d}$ to K sub-clusters. Then, we update the center of the sub-clusters, \eg, $P_{i}=[P_{i}^{1}, ... ,P_{i}^{C}]$, where $P_{i}^{c}=[{p_{c}}_{1},..,{p_{c}}_{K}]$ contains the sub-clusters' center of class $c$ at client $i$. 
    
    To get a more robust and representative sub-clusters, we impose two constraints on the mapping function $Q^{c}$: 
    1) one-hot feature assignment, \eg, $Q^{c} \in \{0,1\}^{K}$ and $\sum_{q_{i}\in Q^{c}} q_{i} \mathbbm{1}=|D^c_i|$ where each feature is  assigned to only one sub-cluster, and 2) equipartition constraint,  $\sum_{q_{i}\in Q^{c}} \mathbbm{1}_{q_{i}==k}=|D^c_i|/K$ where each sub-cluster $k$ holds roughly the same amount of features to avoid degeneration. $Q^{c}$ can be approximated by formulating the above requirement~\cite{transport_relax} as an optimal transport problem presented in Eq.~\ref{subclusters}, which can be solved via fast sinkhorn-knopp algorithm~\cite{sinkhorn}. 
 
    \begin{equation}\label{subclusters}
        Q^{c*} = \text{diag(}\alpha\text{)}(\dfrac{Z_{i}^{c}P_{i}^{c}}{\epsilon}) \text{diag(}\beta\text{)}
    \end{equation}
    
    where $P_{i}^{c} \in \mathbb{R}^{{d}\times {K}}$. $\alpha \in \mathbb{R}^{|D_{i}^{c}|}$, and $\beta \in \mathbb{R}^{K}$ are two re-normalization vectors, computed via sinkhorn-knopp iteration~\cite{sinkhorn}. $\epsilon=0.05$ is the trade-off coefficient between the convergence speed of the approximation and closeness to the optimal solution. For more details, please refer to ~\cite{dnc,rss_prototype}.   

    We update the feature prototype after each training round according to the Eq.~\ref{update_proto}.

    \begin{equation}\label{update_proto}
        \begin{split}
        P^{c}_{i} = (Z_{i}^{c})^{T}(Q^{c*})^{T} \\ 
        P^{c}_{i} = \dfrac{P^{c}_{i}}{N_{k}} \\
        \end{split}
    \end{equation}

    where $N_{k}$ indicates the number of samples assigned to each sub-cluster. With $P_{i}^{c}$ , sub-clusters center for class $c$, we can then construct the $CK$ sub-clusters for each client $i$, \eg, $P_{i}=[P_{i}^{1}, ... , P_{c}^{C}] \in \mathbb{R}^{d \times CK}$. 

    \subsubsection{sub-clusters regularization}
    $P_{i} = P_{i}/||P_{i}||_{2}$ is first normalized to equalize the importance of each class. The clusters' center is then used to regularize the extracted features according to Eq.~\ref{lnpr}. 

    \begin{equation}\label{lnpr}
	\mathcal{L}_{i}^{npr} = \dfrac{1}{|D_{i}|} \sum_{x_{i},y_{i}\in D_{i}} -\text{log }\dfrac{exp(\text{max} (<z_{i},P_{i}^{y_{i}}>)) }{\sum_{j=1}^{j=C} exp(\text{max}(<z_{i},P_{i}^{j}>))},\\
	\end{equation}

    where $<a,b>$ denotes the matrix multiplication between $a$ and $b$.
    
    \subsection{FedNPR}\label{sec3:4}
    FedNPR consists of \textit{local client update} and \textit{server update}:

    \subsubsection{local client update} First, each client updates its local model using the federated model downloaded from the server. Then, to enhance the generalization and the expressiveness of the feature extractor under high inter-client class variation, each client optimizes a linear combination of parametric approach through minimizing balanced softmax (BSM) loss and a non-parametric regularization loss.

    With BSM loss, each client optimizes a debiased classifier, which helps to maintain a common global objective between different clients despite the inter-client class variation. 
 
    NPR loss complements BSM loss, especially when the data is limited, \eg, minority and tail classes. NPR explicitly models and injects class-relevant and informative signals into the feature space through sub-clusters. This enhances the expressivity and generalizability of the extracted features, improving the performance of the model on unseen data.
    
	\myparagraph{Class prior calibration with balanced Softmax}
	We adopt balanced softmax loss~\cite{balanced_softmax} which uses class frequencies as a prior to calibrate the class distributions. Let $\pi_{i} = [\pi^{0}_{i}, ... , \pi^{|class|}_{i}]$ and $\pi^{c}_{i} = |D^c_i|/|D_i|$ be the frequency of class $c$ at client $i$.

    \begin{equation}\label{lsup}
	\mathcal{L}_{i}^{sup} = \dfrac{1}{|D_{i}|} \sum_{x_{i},y_{i}\in D_{i}} -y_{i}\text{log (}\sigma (g_{v}(f_{u}(x_{i})) + \pi_i \text{)}\\
	\end{equation}
	where $\sigma(.)$ is the softmax function.

	\myparagraph{Feature extractor regularization with NPR}
    With NPR as an additional regularization module, each client $i$ minimizes the overall loss in Eq.~\ref{overall} to obtain the local model update $\nabla \phi_{w}^{i}$. 
	\begin{equation}\label{overall}
	\mathcal{L}_{i} = \mathcal{L}_{i}^{sup} + \lambda*\mathcal{L}_{i}^{npr} 
	\end{equation}
	where $\lambda$ is the hyper-parameter to determine the strength of NPR. Discussion on the importance of the two losses is presented in Section~\ref{disc:components}.

    \begin{figure}[t!]
        \centering
        \includegraphics[width=1 \columnwidth]{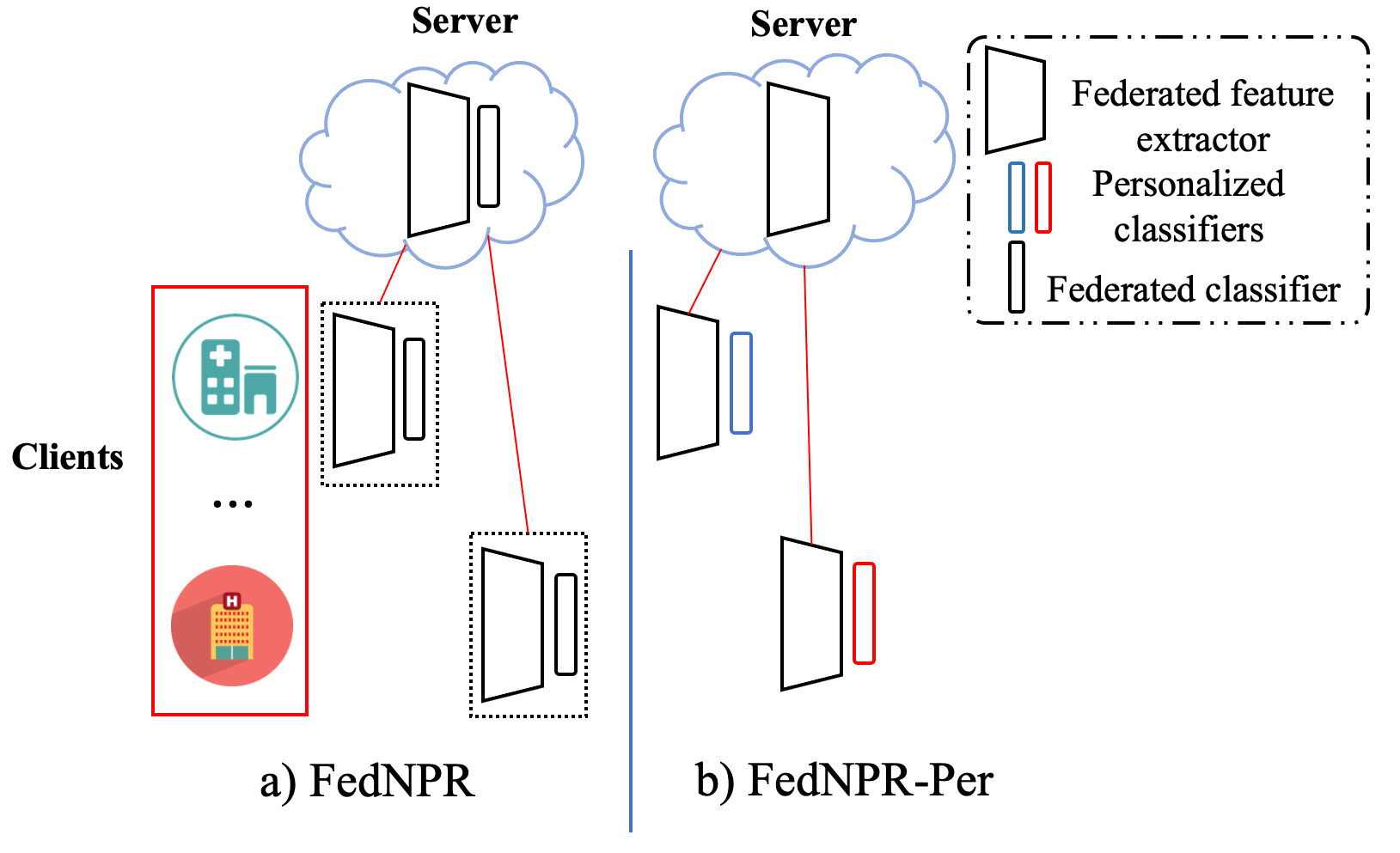}
        \caption{Illustrations of the differences between FedNPR and FedNPR-Pe. FedNPR-Per keeps a personalized classifier head at each client instead of sharing it. Therefore, each client is able to adapt and learn its own classifier which is more aligned with its data characteristics.} \label{fednpr-per}
    \end{figure}
    
	\subsubsection{Federated Model Update}	
	Federated averaging (FedAvg)~\cite{fedavg} is used to update the federated model in the server. Each client $i$ sends its local model update $\nabla \phi_{w}^{i}$ to the server. Each client's importance weight is assigned according to its data amount.
    The server updates the federated model by
	\begin{equation}\label{fedavg}
	\phi_{w} \leftarrow \phi_{w} + \sum_{i=1}^{N} w_{i} \nabla \phi_{w}^{i},
	\end{equation}
	where $w_{i}=|D_i|/\sum_{i=1}^{N} |D_i|$.

	\subsection{FedNPR-Per: Extension to Personalized FL}\label{sec3:7}
    With NPR, the robustness of the feature extractor is improved. From the perspective of transfer learning~\cite{eegdomain,fl_wearable}, a more robust feature extractor is analogous to a better initialized pretrained model. 
    
    In FedNPR-per where each client could benefit more by learning its own classifier head, $g_{v_{i}}$ while collaboratively learning a federated feature extractor $f_{u}$. As presented in Fig.~\ref{fednpr-per}, instead of updating the federated model according to Eq.~\ref{fedavg}, we use Eq.~\ref{fedper}. Consequently, each client $i$ optimizes for its personalized model, $\phi_{i}=\{f_{u}, g_{v_{i}}\}$.

    \begin{equation}\label{fedper}
	f_{u} \leftarrow f_{u} + \sum_{i=1}^{N} w_{i} \nabla f_{u}^{i},
	\end{equation}
	where $w_{i}=|D_i|/\sum_{i=1}^{N} |D_i|$.

	\section{Experiments}\label{evaluation}
	\subsection{Dataset and Preprocessing}
	
	Experiments are conducted on two challenging class imbalance FL tasks:
	\begin{enumerate}
		\item \textbf{Skin lesion classification}. The Fed-ISIC2019~\cite{flamby} dataset contains 23,247 dermoscopy images from six medical sources including eight classes, namely Melanoma, Melanocytic Nevus, Basal Cell Carcinoma, Actinic Keratosis, Benign Keratosis, Dermatofibroma, Vascular Lesion, and Squamous Cell Carcinoma. Each data source is regarded as a separate client. Statistical details of each client's data are presented in Table~\ref{skin_info}, where some clients, \eg, MSK4, ViDIR old, and ViDIR molemax, has missing classes. 
		\textbf{Preprocessing.} Following the recommendations in~\cite{kaggle_preproc}, each dermoscopy image is pre-processed with brightness normalization and color constancy and resized to 224$\times$224 pixels.		
		\item \textbf{Intracranial Hemorrhage (ICH) Classification}. 
		The RSNA-ICH~\cite{rsna} dataset consists of CT images from four different medical sources with five sub-classes including Epidural, Intraparenchymal, Intraventricular, Subarachnoid, and Subdural.
		As data sources are not publicly available, we artificially split the data into two different multi-client settings. \textbf{Preprocessing.} Following \cite{fedirm,dynamic_bank}, 67,969 CT images with single hemorrhage type are selected and resized to 128$\times$128 pixels for training and testing. 
	\end{enumerate}
  
{\renewcommand{\arraystretch}{1.2}
		\begin{table}[t]
			\centering
			\caption{Statistics of different clients from the Fed-ISIC2019 dataset\cite{flamby}, preprocessed and curated from ISIC2019\cite{ham10k,isic2017,bcn2000}, including classes 0 (Melanoma), 1 (Melanocytic Nevus), 2 (Basal Cell Carcinoma), 3 (Actinic Keratosis), 4 (Benign Keratosis), 5 (Dermatofibroma), 6 (Vascular Lesion), and 7 (Squamous Cell Carcinoma).}\label{skin_info}
			\begin{tabular}{c|c|c|c|c|c}
				\hline
				\multirow{3}{*}{Source}  & \multicolumn{4}{c|}{ \# Images per class} & \# Images\\ \cline{2-6}
				& 0    & 1    & 2    & 3    & Train \\ \cline{6-6}
				& 4    & 5    & 6    & 7    & Test \\ \hline
				\multirow{2}{*}{Rosendahl}     & 342  & 803  & 296  & 109  & 1807 \\ \cline{6-6} 
				& 490  & 30   & 3    & 18   & 452     \\ \hline
				\multirow{2}{*}{BCN}           & 2857 & 4206 & 2809 & 737  & 9930 \\ \cline{6-6} 
				& 1138 & 124  & 111  & 431  & 2483     \\ \hline
				\multirow{2}{*}{MSK4}          & 215  & 415  & 0    & 0    & 655 \\ \cline{6-6} 
				& 189  & 0    & 0    & 0    & 164     \\ \hline
				\multirow{2}{*}{VIDIR Modern}  & 680  & 1832 & 211  & 21   & 2691 \\ \cline{6-6} 
				& 475  & 51   & 82   & 11   & 672     \\ \hline
				\multirow{2}{*}{VIDIR Old}     & 67   & 350  & 5    & 0    & 351 \\ \cline{6-6} 
				& 10   & 4    & 3    & 0    & 88   \\ \hline
				\multirow{2}{*}{VIDIR Molemax} & 24   & 3720 & 2    & 0    & 3163 \\ \cline{6-6} 
				& 124  & 30   & 43   & 0    & 791     \\ \hline
			\end{tabular}
		\end{table}
	}
	{\renewcommand{\arraystretch}{1.2}
		\begin{table}[t]
			\centering
			\caption{Statistics of different clients for federated intracranial CT hemorrhage detection~\cite{rsna} under the 10-client setting with severely imbalance inter-client class variations and missing classes, including classes 0 (Epidural), 1 (Intraparenchymal), 2 (Intraventricular), 3 (Subarachnoid), and 4 (Subdural).}\label{brain_info_split2}
			\begin{tabular}{c|c|c|c|c|c|c|c}
				\hline
				\multirow{2}{*}{Site}    & \multicolumn{5}{c|}{ \# Images per class} & \multicolumn{2}{c}{ \# Images} \\ \cline{2-8}
				& 0   & 1    & 2    & 3    & 4    & Train & Test \\ \hline
				\#1  & 0   & 0    & 1112 & 2245 & 4033 & 7390  & 1848     \\ \hline
				\#2  & 88  & 5522 & 0    & 1926 & 0    & 7536  & 1884     \\ \hline
				\#3  & 89  & 1122 & 2339 & 0    & 7627 & 11177 & 2795     \\ \hline
				\#4  & 0   & 2430 & 0    & 3574 & 2944 & 8948  & 2237     \\ \hline
				\#5  & 132 & 518  & 0    & 930  & 1828 & 852   & 2744     \\ \hline
				\#6  & 269 & 0    & 0    & 1034 & 1834 & 3137  & 785     \\ \hline
				\#7  & 538 & 928  & 2662 & 0    & 0    & 1033  & 2616     \\ \hline
				\#8  & 9   & 502  & 0    & 0    & 2869 & 3380  & 845     \\ \hline
				\#9  & 69  & 62   & 0    & 1635 & 957  & 2723  & 681     \\ \hline
				\#10 & 4   & 62   & 1017 & 429  & 1033 & 2545  & 637     \\ \hline
			\end{tabular}
		\end{table}
	}

	\subsection{Evaluation}
	
	{\renewcommand{\arraystretch}{1.3}
		\begin{table*}[t]
			\centering
			\caption{
				Quantitative results of different learning frameworks for federated class imbalance skin lesion and intracranial hemorrhages classification. Each learning framework is trained under five different seeds and both the average performance and standard deviation are reported. The best results are marked in \textbf{bold}. 
			}
			\label{results_full}
               \fontsize{10pt}{10pt}\selectfont
			\begin{tabular}{l|cc|cc}
				\hline
				\multirow{2}{*}{Method} &  \multicolumn{2}{c|}{Fed-ISIC 2019~\cite{flamby} (\%)}   
				&  \multicolumn{2}{c}{RSNA-ICH~\cite{rsna} (\%)}   \\ \cline{2-5} 
				& bACC & bAUC & bACC & bAUC  \\ \cline{1-5}
				Local Learning                          & 70.0$\pm$0.7 & 88.3$\pm$1.2 & 60.7$\pm$0.8 & 89$\pm$0.5  \\ 
				FedAvg~\cite{fedavg}                    & 69.5$\pm$0.6 & 91.3$\pm$0.7 & 59.8$\pm$0.4   & 84.8$\pm$0.2  \\ 
				FedProx (MLSys20)~\cite{fedprox}        & 70.0$\pm$1.1 & 90.2$\pm$0.2 & 64.8$\pm$0.5 & 89.4$\pm$0.4  \\ 
				MOON( CVPR21)~\cite{moon}               & 69.5$\pm$1.0 & 90.5$\pm$0.8 & 61.5$\pm$0.6 & 88.4$\pm$0.6 \\ 
				CReRF (IJCAI22)~\cite{crerf}            & 67.6$\pm$1.3 & 89.6$\pm$1.2 & 60.8$\pm$0.2 & 89$\pm$0.4  \\ 
				FedRS (KDD21)~\cite{fedrs}              & 70.2$\pm$2.2 & 90.3$\pm$0.3 & 62.8$\pm$0.7 & 83.6$\pm$0.3  \\ 
				FedLC (ICML22)~\cite{fedlc}             & 67.8$\pm$0.9 & 89.3$\pm$0.9 & 63.7$\pm$0.7 & 86.1$\pm$0.4 \\ 
				BalanceFL (IPSN22)~\cite{balancefl}     & 66.4$\pm$1.0 & 88.2$\pm$0.4 & 58.2$\pm$0.7 & 87.6$\pm$0.2 \\ 
				FedREP (ICML21)~\cite{fedrep}           & 69.8$\pm$1.3 & 88.9$\pm$0.5 & 63.4$\pm$0.5 & 89.3$\pm$0.4\\ 
                    FedRod (ICLR22)~\cite{fedrod}         & 66.1$\pm$0.8 & 90.5$\pm$1.5 & 64.7$\pm$0.5 & 89.8$\pm$0.4 \\
				FedBABU (ICLR22)~\cite{fedbabu}         & 72.1$\pm$1.6 & 90.5$\pm$1.5 & 64.4$\pm$0.5 & 89$\pm$0.5 \\ \hline
                    FedNPR (ours) & 72.9$\pm$0.6 & 91.5$\pm$0.7 & 65.4$\pm$0.8 & 87.5$\pm$0.5 \\
                    FedNPR-Per (ours)                              & \textbf{76.2}$\pm$0.9 & \textbf{91.8}$\pm$0.5 & \textbf{72.1}$\pm$0.3 & \textbf{90.7}$\pm$0.5
                \\\hline
			\end{tabular}
		\end{table*}
	}

	For each dataset/source, we use 80\% for training and 20\% for testing while preserving the same class ratio. The average performance and standard deviation of different learning frameworks through \textbf{five-fold} cross validation are reported for comparison. 
	
	\myparagraph{Metric.}
	Balanced accuracy (bACC), average per class ACC, balanced area under the curve (bAUC), and average per class AUC are jointly used for evaluation.
 
	\myparagraph{Settings.}
	We evaluate model performance separately on each client's test set, $D^{k}_{test}$, and measured by average bACC and bAUC over clients, namely $\frac{1}{K}\sum_{k=1}^{K} \text{bACC}({D^{k}_{test}}$) and $\frac{1}{K}\sum_{k=1}^{K} \text{bAUC}({D^{k}_{test}}$). 
	
	\subsection{Implementation Details}
	\myparagraph{Network architectures.} Following~\cite{skin1,skin2}, EfficientNet-B0~\cite{efficientnet} is used as the baseline model architecture. We use EfficientNet-B0 because it is empirically shown to be better at capturing more generalizable features for imbalance medical image classification task under the centralized setting~\cite{anal_isic_mia}. 

	\myparagraph{Comparison methods.} We compare FedNPR and FedNPR-Per with recent state-of-the arts federated approaches from both 1) single-model FL and 2) personalized FL. 
    
    From single-model FL, we include 1) local learning, where each client trains a model individually, 2) FedAvg~\cite{fedavg} as a baseline comparison, and representative approaches including FedProx~\cite{fedprox}, MOON~\cite{moon}, CReRF~\cite{crerf}, and the most-recent state-of-the-art approaches such as BalanceFL~\cite{balancefl}, FedRS~\cite{fedrs}, and FedLC~\cite{fedlc}. 

    From personalized FL approaches, we present the recent state-of-the-art approaches such as FedRep~\cite{fedrep}, FedBABU~\cite{fedbabu}, and FedRod~\cite{fedrod}.

    For FedNPR, we perform hyperparameter grid search on $K\in \{1,2,4,8\}$ and $\lambda \in \{0.01, 0.05, 0.1, 1\}$ and found the best hyperparameters to be $K=4$ and $\lambda=0.1$ for federated skin lesion classification and $K=2$ and $\lambda=0.05$ for ICH classification. 
    We provide detailed analysis on the importance of the hyperparameter in Section~\ref{discussion}.
	
	\myparagraph{Training Details}. EfficientNet-B0 is initialized with the pre-trained weights from ImageNet and trained for 80 federated rounds using an Adam optimizer~\cite{adam} with a learning rate of 1e-3, a weight decay of 5e-4, and a batch size of 64. We apply learning rate decay with a factor of 0.1 at round 60 and 70. In synchronous federated training, each client $i$ updates the modal locally for one epoch and sends local model updates to the server at every federated round. Training images are augmented by random rotation, horizontal and vertical flipping, adding gaussian blur, and applying normalization. Testing images are normalized according to the training statistics. 
    
    For a fair comparison, balanced softmax loss (BSM) \cite{balanced_softmax} is introduced to optimize all learning frameworks as it is shown to work better than regular cross entropy and focal loss~\cite{focal_loss} for class-imbalanced learning. 

	\subsection{Results on Skin Lesion Classification}
	
	\subsubsection{Experiment Settings}
	
	Fed-ISIC2019~\cite{flamby} is divided into six clients according to data sources: Rosendahl, BCN, MSK4, VIDIR Modern, VIDIR Old, and VIDIR Molemax respectively. As summarized in Table.~\ref{skin_info}, clients vary significantly in data amounts and class distributions, and there exist missing classes in MSK4, VIDIR Old, and VIDIR Molemax.
		
	\subsubsection{Comparison of Various Learning Frameworks}
	Table~\ref{results_full} highlights the impact of severe inter-client class variation on the performance of different federated learning (FL) frameworks, including local learning (LL).  
	Compared to LL, the baseline FedAvg model achieves a 2\% improvement in bAUC, but its bACC is 0.5\% lower. Notably, LL outperforms most existing FL approaches, except for FedProx, FedRS, and FedBABU. 
	The results suggest that in scenario where inter-client class variation is severe, collaborative learning may not be as effective as local learning, which could discourage some clients from participating. This highlights the importance of developing FL frameworks that can effectively handle class variation across clients.  
	 
	Compared to existing federated learning (FL) approaches designed to handle inter-client variation, FedAvg demonstrates comparable performance and even outperforms certain methods, such as CReRF, FedLC, BalanceFL, which focus on calibrating the federated model for a more balanced class distribution, and FedREP, FedRod, personalized FL frameworks that optimize a shared federated model and its personalized classifier head. The observation indicates that these methods may not be effective in scenarios where the degree of class imbalance and inter-client class variation is too high.
 
	Based on the evaluation in Fed-ISIC, approaches that outperform FedAvg tend to impose less restrictive regularization. FedProx, for instance, improves upon FedAvg's performance by 0.5\% in bACC thanks to its simple and general variation reduction technique. FedRS slightly outperforms FedAvg in bACC by an average of 0.7\% since it diminishes the weight updates on missing classes, causing clients to focus solely on existing classes. It is worth noting that CReRF performs poorly compared to other approaches, as it relies on the generated features on the server to calibrate the classifier. When clients' local updates diverge, the quality of the generated features would be negatively affected. 
    FedBABU benefits from having a more consistent target across clients, such as a frozen classifier during training, which results in an increase of 2.1\% in bACC compared to FedAvg. However, these methods achieve lower bAUC compared to FedAvg, likely because directly calibrating the classifier at each client according to its distribution without regularization may distort the model's decision boundaries.
	
    FedNPR addresses the aforementioned limitations with NPR module. FedNPR outperforms the state-of-the-art federated and local learning approaches by an increase of 0.8\% and 0.3\% in bACC and bAUC. 
    NPR module explicitly captures structural information and useful signals in the data and model them in the feature space as anchors. With such anchors used to regularize the local model update, the extracted features' expressivity, compactness, and discriminability are further enhanced. When the feature extractor can extract generalizable features across clients, it reduces the model update divergence between clients. 

        {\renewcommand{\arraystretch}{1.3}
		\begin{table*}[t]
			\centering
			\caption{
				Ablation studies on various components in FedNPR-Per: 1) global supervision (via BSM loss) and
                2) non-parametric regularization with single sub-cluster or multiple sub-clusters. Each variation is trained under five different seeds and both the average performance and standard deviation are reported. The best results are marked in \textbf{bold}.
			}
			\label{npr:components}
               \fontsize{10pt}{10pt}\selectfont
			\begin{tabular}{c|cc|cc|cc}
				\hline
				\multicolumn{3}{c|}{Components} &  \multicolumn{2}{c|}{\multirow{2}{*}{Fed-ISIC 2019~\cite{flamby} (\%)}}   
				&  \multicolumn{2}{c}{\multirow{2}{*}{RSNA-ICH~\cite{rsna} (\%)}}   \\ \cline{1-3}
                \multirow{2}{*}{global supervision} & \multicolumn{2}{c|}{Non-parametric regularization (NPR)} &  &  &  &  \\ \cline{2-7}
				& single sub-cluster & multiple sub-clusters & bACC & bAUC & bACC & bAUC  \\ \hline
                 \multirow{2}{*}{$\times$} & \checkmark & & 57.4$\pm$1.1 & 83.2$\pm$1.7 & 58.5$\pm$2.9 &82.0$\pm$1.4 \\ 
                 & & \checkmark & 60.4$\pm$2.3 & 83.5$\pm$1.3 & 60.7$\pm$1.8 & 83.7$\pm$1.6 \\ \hline
                \multirow{2}{*}{\checkmark} & \checkmark & & 73.3$\pm$0.1 & 91.4$\pm$0.9 & 70.8$\pm$0.6 &88.8$\pm$0.3 \\ 
                 & & \checkmark & \textbf{76.2$\pm$0.9} & \textbf{91.8$\pm$0.5} & \textbf{72.1$\pm$0.3} & \textbf{90.7$\pm$0.5} \\ \hline
			\end{tabular}
		\end{table*}
	}

    With the more generalizable federated feature extractor, each client can optimize its personalized classifier head without drastically changing the feature extractor during local updates. We then extend FedNPR into personalized FL as FedNPR-Per by allowing each client to further benefit by optimizing its personalized classifier head. This enables each client to adapt the federated model to its unique data characteristics, leading to better performance and customization. FedNPR-Per improves upon FedNPR by 3.3\% in bACC and 0.3\% in bAUC. The results demonstrate that a more generalizable federated feature extractor enables different clients to collaborate more effectively and adapt the federated model with ease to their unique data characteristics.

    \subsection{Results on Intracranial Hemorrhage Classification}
	
    \subsubsection{Experimental Settings.}
    Though the RSNA-ICH~\cite{rsna} dataset was collected from four different medical sources, the data source of each image is unknown. Therefore, following \cite{moon, crerf}, we use a Dirichlet distribution for data partitioning with cross-client class variations. Dirichlet distribution is generated according to a hyper-parameter $\alpha$, where a higher $\alpha$ would lead to a more balanced distribution. We simulate imbalance class distribution, \eg, inter-client class variations with 10-clients for our study. We use five different Dirichlet distributions for classes, \ie, Subdural with $\alpha=50$, Subarachnoid with $\alpha=30$,  Intraventricular with $\alpha=10$, Intraparenchymal with $\alpha=5$, and Epidural with $\alpha=0.5$ respectively. To simulate missing classes, we randomly remove classes at each client with a probability of $0.3$. Statistical details are stated in Table~\ref{brain_info_split2}.

	\subsubsection{Comparison of Various Learning Frameworks}
	The trend in ICH classification mirrors the skin lesion classification, where local learning surpasses vanilla FedAvg by a significant margin, with improvements of 0.9\% in bACC and 5.2\% in bAUC, highlighting the severity of inter-client class variations. We observe that most of the existing state-of-the-art FL techniques outperform FedAvg in both bACC and bAUC. The improvements indicate that the state-of-the-art FL approaches are able to operate well for ICH classification. It is noteworthy that existing state-of-the-art FL experiments are commonly conducted on an artificially split dataset with a Dirichlet distribution, similar to the setting employed for ICH classification.

    FedProx, a simple yet effective method, outperforms existing FL approaches with 64.8\% bACC and 89.4\% bAUC, surpassing FedAvg by 5\% and 4.6\% in bACC and bAUC, respectively. This highlights the effectiveness of minimizing model update variances in improving the generalization of the federated model. In general, personalized FL methods, such as FedREP, FedRod, and FedBABU, exhibit better performance than single model FL, with ranges of 63.4-64.4\% in bACC and 89.0-89.8\% in bAUC, compared to 58.2-64.8\% in bACC and 86.1-89.4\% in bAUC. By adopting personalized FL, each client can derive greater benefits, particularly when local model updates exhibit minimal divergence.

    FedNPR significantly outperforms FedAvg by 5.6\% in bACC and 2.7\% in bAUC, demonstrating the advantage of having a more generalizable feature extractor. Furthermore, FedNPR-Per, which extends FedNPR with personalized classifier heads, further improves the performance of FedNPR by 6.7\% and 3.2\% in bACC and bAUC, respectively. This leads to a total improvement of 7.3\% and 0.9\% in bACC and bAUC, respectively, over the state-of-the-art FL approaches. This result highlights the effectiveness of combining a generalizable feature extractor with personalized classifier heads in FL settings.

	\section{Ablation Study}\label{discussion}
    In this section, we investigate the role of hyperparameters and the components of NPR under the federated setting. First, we study the roles of NPR in FedNPR-Per to further highlight the differences between each component. Then, we explore how NPR can enhance existing FL techniques, such as FedAvg and FedProx, and discuss potential challenges and limitations.

    \subsection{Components of FedNPR-Per}\label{disc:components}   
    We divide the components of FedNPR-Per according to the loss it optimizes in Eq~\ref{overall}, \eg, the BSM loss, as a global supervision between various clients, and the NPR loss. For NPR loss, we vary $K$, the number of sub-cluster(s) assigned to each client, \eg,  $K=1$ and $K>1$. 

    Table~\ref{npr:components} shows consistent results on both skin lesion classification and ICH identification. First, it is essential to have a common objective across clients, \eg, the global supervision. 
    Without the same objective, it is difficult to prevent the each client's model update from diverging. 
    Second, multiple sub-clusters is better than single-cluster NPR. Multiple sub-clusters can better model and capture class-specific characteristic in the feature space to guide and regularize the feature extractor. 
    FedNPR-Per with global supervision and multiple sub-clusters outperform the FedNPR-per without global supervision and single sub-cluster by 18.6\% and 8.6\% in bACC and bAUC respectively for skin-lesion classification, and 13.6\% and 8.7\% in bACC and bAUC respectively for ICH identification. 

    {\renewcommand{\arraystretch}{1}
		\begin{table}[t]
			\centering
			\caption{
				Ablation studies of FedNPR-Per on the number of clusters $K$. Each variation is trained under five different seeds and both the average performance and standard deviation are reported. The best results are marked in \textbf{bold}. 
			}\label{ablation_clusters}
                \fontsize{9.5pt}{9.5pt}\selectfont
			\begin{tabular}{c|cc|cc}
				\hline
				\multirow{3}{*}{$K$}  
				& \multicolumn{2}{c|}{Fed-ISIC 2019~\cite{flamby} (\%)}  & \multicolumn{2}{c}{RSNA-ICH~\cite{rsna} (\%)} \\ \cline{2-5} 
                & \multicolumn{2}{c|}{bACC} & \multicolumn{2}{c}{bACC}\\
                & \multicolumn{2}{c|}{bAUC} & \multicolumn{2}{c}{bAUC}\\ \hline 
                \multirow{2}{*}{1} & \multicolumn{2}{c|}{73.3$\pm$0.1}& \multicolumn{2}{c}{70.8$\pm$0.6} \\ 
                &  \multicolumn{2}{c|}{91.4$\pm$0.9}& \multicolumn{2}{c}{88.8$\pm$0.3} \\ \hline 
                \multirow{2}{*}{2} & \multicolumn{2}{c|}{74.8$\pm$0.9}& \multicolumn{2}{c}{\textbf{72.1$\pm$0.3}} \\ 
                &  \multicolumn{2}{c|}{91.5$\pm$0.5}& \multicolumn{2}{c}{\textbf{90.7$\pm$0.5}} \\ \hline 
                \multirow{2}{*}{4} & \multicolumn{2}{c|}{\textbf{76.2$\pm$0.9}}& \multicolumn{2}{c}{70.6$\pm$0.3} \\ 
                &  \multicolumn{2}{c|}{\textbf{91.8$\pm$0.5}}& \multicolumn{2}{c}{89.4$\pm$0.7} \\ \hline 
                \multirow{2}{*}{8} & \multicolumn{2}{c|}{74.5$\pm$0.9}& \multicolumn{2}{c}{70.5$\pm$0.9} \\ 
                &  \multicolumn{2}{c|}{91.4$\pm$0.5}& \multicolumn{2}{c}{88.7$\pm$0.7} \\ \hline 
			\end{tabular}
		\end{table}
	}
 
	\subsection{NPR - number of clusters}\label{npr:cluster}
    By utilizing multiple sub-clusters, each client can better model its class characteristic in the feature space, which helps regularize the feature extractor. We identified the optimal value of $K=4$ and $K=2$ for skin lesion classification and ICH identification, respectively. From Table~\ref{ablation_clusters}, the corresponding bACC and bAUC with the optimum $K$ value are 76.2\% and 91.8\% and 72.1\% and 90.7\% for skin lesion classification and ICH identification, respectively. As the value of $K$ increases further, the performance of FedNPR-Per drops, suggesting that increasing the number of sub-clusters may reduce the compactness of each class' specific features, thereby limiting the inter-class feature separability.
    
	\subsection{NPR - lambda value}\label{npr:lambda}
    We found that setting $\lambda=0.1$ and $\lambda=0.05$ results in the best performance for skin-lesion classification and ICH identification, respectively. The results in Table~\ref{ablation_lam} show that setting $\lambda=1$ does not lead to the best-performing model. This indicates two things: 1) minimizing the global supervision loss, which is a common objective between different clients, is still essential, further supporting the finding in Section~\ref{disc:components}, and 2) NPR's role is primarily limited to a regularization module. In other words, relying solely on a non-parametric approach to learn a robust and more generalizable feature extractor is challenging. 

 {\renewcommand{\arraystretch}{1}
		\begin{table}[t]
			\centering
			\caption{
				Studies of combining NPR with various representative FL techniques. The performance improvement compared to the baseline technique in absence of NPR is presented in bracket with blue with \textcolor{blue}{$\uparrow$}, best viewed in color.
			}\label{npr_various}
                \fontsize{9pt}{9pt}\selectfont
			\begin{tabular}{c|cc|cc}
				\hline
				\multirow{3}{*}{FL Techniques}  
				& \multicolumn{2}{c|}{Fed-ISIC 2019~\cite{flamby}(\%)}  & \multicolumn{2}{c}{RSNA-ICH~\cite{rsna}(\%)} \\ \cline{2-5} 
                & \multicolumn{2}{c|}{bACC} & \multicolumn{2}{c}{bACC}\\
                & \multicolumn{2}{c|}{bAUC} & \multicolumn{2}{c}{bAUC}\\ \hline 
                \multirow{2}{*}{FedAvg+NPR} & \multicolumn{2}{c|}{72.9 (\textcolor{blue}{$\uparrow$ 3.4})}& \multicolumn{2}{c}{65.4 (\textcolor{blue}{$\uparrow$ 5.6})} \\ 
                &  \multicolumn{2}{c|}{91.5 (\textcolor{blue}{$\uparrow$ 0.2})}& \multicolumn{2}{c}{87.5 (\textcolor{blue}{$\uparrow$ 2.7})} \\ \hline 
                \multirow{2}{*}{FedBABU+NPR} &  \multicolumn{2}{c|}{73.7 (\textcolor{blue}{$\uparrow$ 1.6})}& \multicolumn{2}{c}{71.7 (\textcolor{blue}{$\uparrow$ 7.3})} \\ 
                &  \multicolumn{2}{c|}{90.8 (\textcolor{blue}{$\uparrow$ 0.3})}& \multicolumn{2}{c}{90.3 (\textcolor{blue}{$\uparrow$ 1.3})} \\ \hline 
                \multirow{2}{*}{FedREP+NPR} &  \multicolumn{2}{c|}{70.8 (\textcolor{blue}{$\uparrow$ 1.0})}& \multicolumn{2}{c}{70.4 (\textcolor{blue}{$\uparrow$ 7.0})} \\
                &  \multicolumn{2}{c|}{90.4 (\textcolor{blue}{$\uparrow$ 1.5})}& \multicolumn{2}{c}{90.2 (\textcolor{blue}{$\uparrow$ 0.9})} \\ \hline 
                \multirow{2}{*}{FedROD+NPR} &  \multicolumn{2}{c|}{67.3 (\textcolor{blue}{$\uparrow$ 1.2})}& \multicolumn{2}{c}{67.1 (\textcolor{blue}{$\uparrow$ 2.4})} \\
                &  \multicolumn{2}{c|}{91.4 (\textcolor{blue}{$\uparrow$ 0.9})}& \multicolumn{2}{c}{91.9 (\textcolor{blue}{$\uparrow$ 2.1})} \\ \hline 
			\end{tabular}
		\end{table}
	}
    
    \subsection{Can we combine NPR with Existing FL Techniques?}\label{npr:fl}
    The NPR module can be independently added to regularize the feature extractor during the local update of each client, thereby improving the generalization of the learned feature extractor, independent of the employed federated approaches. 
    Following the results on \ref{npr:lambda}, we set $\lambda=0.1$ and $\lambda=0.05$ for experiments on skin lesion and ICH identification respectively. 
    
    Experimental results in Table.~\ref{npr_various} indicate that adding NPR consistently enhances the model performance learned through various FL approaches between 1-3.4\% and 0.3-5.6\% for bACC and bAUC respectively on skin-lesion classification and 0.2-7.3\% and 0.9-2.7\% for bACC and bAUC on ICH identification. These results suggest that NPR effectively regularizes the feature extractor and improves the generalization of the federated model, leading to better performance on both tasks.

    {\renewcommand{\arraystretch}{1}
		\begin{table}[t]
			\centering
			\caption{
				Ablation studies of the regularization weight $\lambda$ in FedNPR-Per. Each variation is trained under five different seeds and both the average performance and standard deviation are reported. The best results are marked in \textbf{bold}. 
			}\label{ablation_lam}
                \fontsize{9.5pt}{9.5pt}\selectfont
			\begin{tabular}{c|cc|cc}
				\hline
				\multirow{3}{*}{$\lambda$}  
				& \multicolumn{2}{c|}{Fed-ISIC 2019~\cite{flamby} (\%)}  & \multicolumn{2}{c}{RSNA-ICH~\cite{rsna} (\%)} \\ \cline{2-5} 
                & \multicolumn{2}{c|}{bACC} & \multicolumn{2}{c}{bACC}\\
                & \multicolumn{2}{c|}{bAUC} & \multicolumn{2}{c}{bAUC}\\ \hline 
                \multirow{2}{*}{0.01} & \multicolumn{2}{c|}{71.8$\pm$0.6}& \multicolumn{2}{c}{70.5$\pm$2.1} \\ 
                &  \multicolumn{2}{c|}{91.9$\pm$0.3}& \multicolumn{2}{c}{89.4$\pm$0.5} \\ \hline 
                \multirow{2}{*}{0.05} & \multicolumn{2}{c|}{72.7$\pm$0.8}& \multicolumn{2}{c}{\textbf{72.1$\pm$0.3}} \\ 
                &  \multicolumn{2}{c|}{90.9$\pm$0.3}& \multicolumn{2}{c}{\textbf{90.7$\pm$0.5}} \\ \hline 
                \multirow{2}{*}{0.1} & \multicolumn{2}{c|}{\textbf{76.2$\pm$0.9}}& \multicolumn{2}{c}{70.5$\pm$0.5} \\ 
                &  \multicolumn{2}{c|}{\textbf{91.8$\pm$0.5}}& \multicolumn{2}{c}{90.5$\pm$0.4} \\ \hline 
                \multirow{2}{*}{1} & \multicolumn{2}{c|}{73.3$\pm$0.6}& \multicolumn{2}{c}{69.2$\pm$1.1} \\ 
                &  \multicolumn{2}{c|}{92.1$\pm$0.5}& \multicolumn{2}{c}{88.5$\pm$1.3} \\ \hline 
			\end{tabular}
		\end{table}
	}
    
    \subsection{Limitations of FedNPR}
    While NPR module consistently improves the performance of various FL approaches, it relies on: 1) global supervision loss, and 2) representative clusters.
    
    On its own, NPR is not able to explicitly model the relationship between different classes. NPR loss is designed to increase the compactness of the features belonging to a certain class. 
    As a result, in Table.~\ref{npr:components}, FL without global supervision, the performance of federated model learned with only NPR underperforms by 11.4\%-15.9\% in bACC and 6.8\%-8.2\% in bAUC. 
    The observation is aligned with the results in Table.~\ref{ablation_lam}, where the optimal importance weight of NPR, $\lambda$, is around 0.05-0.1. 
    
    In practice, when handling class imbalance, the ideal number of clusters $K$ used for different client and different classes may vary. 
    For instance, when the number of data for a particular rare class/disease is limited, it is less beneficial to use high number of cluster $K$ as each cluster may only have one sample or even no sample. 
    On the other hand, when the data of a particular class is more abundant, it is more beneficial to use a higher number of cluster $K$ to ensure that the intra-class variations are well captured in the feature space. 
    However, to simplify the optimization process and hyperparameter tuning, in this work, we use the same $K$ sub-clusters per class for all clients.
    
    The role of NPR is to further enhance the discriminativeness and compactness of the extracted features. Therefore, when applying NPR to existing FL approaches, it is essential to not only ensure that the global supervision loss aligns objectives across clients but also ensure that the sub-clusters are representative enough by carefully tuning the hyperparameters of NPR. 

	\section{Conclusion}\label{conclusion}
	This paper highlights a commonly encountered yet challenging problem under the federated setting, namely, severe inter-client class variations.
    Severe inter-client class variations can lead to divergent model updates during the local update phase of each training round, negatively impacting the federated model.
    We address the issue by integrating non-parametric regularization (NPR) into federated learning, \eg, FedNPR and FedNPR-Per as both single-model FL and personalized FL. 
    NPR enhances the robustness and expressivity of the extracted feature by directly modeling different class' characteristic in the feature space in a form of $K$ sub-clusters and using them as an anchor to guide the feature extractor.
	Extensive evaluation on two challenging multi-source class imbalance federated learning on skin lesion and intracranial hemorrhage classification shows that FedNPR and FedNPR-Per consistently outperform existing state-of-the-art FL approaches.
    Additionally, we also show that the NPR module can consistently improve the performance of existing state-of-the-art federated techniques. 
    Therefore, we consider the NPR module to be a valuable and effective tool in addressing the challenges of inter-client class variation in federated learning.

\end{document}